%% file: paper_v1.0.tex
\documentclass[10pt,twocolumn,letterpaper]{article}

\usepackage{wacv}
\usepackage{times}
\usepackage{epsfig}
\usepackage{graphicx}
\usepackage{amssymb,amsthm,amsmath}
\usepackage{bm}
\usepackage{booktabs} 
\usepackage[caption=false]{subfig}
\usepackage{makecell}
\usepackage{multirow}
\newcommand{\tabincell}[2]{\begin{tabular}{@{}#1@{}}#2\end{tabular}}

\usepackage[pagebackref=true,breaklinks=true,letterpaper=true,colorlinks,bookmarks=false]{hyperref}

\wacvfinalcopy 

\def\wacvPaperID{473} 

\begin{document}
	
	\title{Offset Calibration for Appearance-Based Gaze Estimation via Gaze Decomposition}
	
	\author{Zhaokang Chen \hspace{2cm} Bertram E. Shi \\
		The Hong Kong University of Science and Technology\\
		{\tt\small \hspace{-1.0cm} zchenbc@connect.ust.hk \hspace{1.2cm} eebert@ust.hk}
	}
	\maketitle
	
	\begin{abstract}
		Appearance-based gaze estimation provides relatively unconstrained gaze tracking. However, subject-indepen\-dent models achieve limited accuracy partly due to individual variations. To improve estimation, we propose a novel gaze decomposition method and a single gaze point calibration method, motivated by our finding that the inter-subject squared bias exceeds the intra-subject variance for a subject-independent estimator. We decompose the gaze angle into a subject-dependent bias term and a subject-independent term between the gaze angle and the bias. The subject-independent term is estimated by a deep convolutional network. For calibration-free tracking, we set the subject-dependent bias term to zero. For single gaze point calibration, we estimate the bias from a few images taken as the subject gazes at a point. Experiments on three datasets indicate that as a calibration-free estimator, the proposed method outperforms the state-of-the-art methods by up to $10.0\%$. The proposed calibration method is robust and reduces estimation error significantly (up to $35.6\%$), achieving state-of-the-art performance for appearance-based eye trackers with calibration.
	\end{abstract}
	
	\section{Introduction}
	As an important cue about people's intent, eye gaze has been used in many promising real-world applications, such as human-computer interfaces~\cite{menges2017gazetheweb,pi2017probabilistic}, human-robot interaction~\cite{huang2016anticipatory}, virtual reality~\cite{outram2018anyorbit,patney2016towards}, social behavioral analysis~\cite{hoppe2018eye} and health care~\cite{grillini2018towards}. These successes have lead to gaze tracking attracting more and more attention.
	
	To date, most eye trackers have relied upon active illumination, e.g. infrared illumination used in pupil center corneal reflections (PCCR). While these provide high accuracy, they also place strong constraints on users' head movements. Accuracy rapidly degrades as the head pose changes. Nonetheless, these techniques are commonly used in laboratory settings where high accuracy is required. Researchers have proposed many novel methods to alleviate constraints on head movement, which will enable more real-world applications that require unconstrained gaze estimation in more flexible environments, e.g.~\cite{brau2018multiple,chong2018connecting,fuhl2018cbf,pi2019task,wang2018slam}. However, a common disadvantage of active illumination approaches is that they are also relatively costly, as they rely upon custom hardware to provide the required illumination.
	
	Appearance-based gaze estimation estimates the gaze directions based on RGB images. It is attracting more and more attention because it provides relatively unconstrained gaze tracking and requires only commonly available off-the-shelf cameras. However, obtaining high accuracy is very challenging due to large variability caused by factors such as differences in individual appearance, head pose, and illumination~\cite{zhang2019mpiigaze}. The application of deep convolutional neural networks (CNNs) to this problem has reduced estimation error significantly~\cite{zhang2015appearance}. There are a large number of real and synthetic datasets covering a wide range of these variations
	\cite{fischer2018rt,funes2014eyediap,krafka2016eye,shrivastava2017learning,smith2013gaze,sugano2014learning,Wang_2018_CVPR,wood2016learning,zhang2015appearance}. Using these datasets, it has been shown that deep CNNs can learn to compensate for the variability \cite{chen2018appearance,cheng2018appearance,deng2017monocular,krafka2016eye,lian2018multiview,ranjan2018light,zhang2017s}.
	
	Unfortunately, the estimation error of subject-indepen\-dent appearance-based methods is still higher than that acheivable using active illumination, e.g. $\sim5^\circ$ vs $\sim1^\circ$. Thus, further work must be done to reduce this error. 
	
	One way to further reduce estimation error is through personal calibration. PCCR-based eye trackers typically require an initial multiple gaze point calibration step before usage. The user must gaze at a number (typically nine points on a three by three gird) of targets sequentially ~\cite{guestrin2006general}. This enables  subject-specific parameters of a geometric 3D eye model to be estimated. A similar calibration procedure has been proposed for appearance-based methods, where some parameters of the estimator are fine-tuned based on the calibration data ~\cite{krafka2016eye,linden2018appearance,liu2018differential}. However,  multiple gaze point calibration is time-consuming and may not be applicable in some situations, e.g. in screen-free applications, it is difficult to provide multiple targets for calibration.
	
	In this article, we propose a gaze decomposition method for appearance-based gaze estimation, and a single gaze point calibration method that requires the user to gaze only at a single point. This is more widely applicable than multiple gaze point calibration e.g. for screen-free applications, since the camera, which should always be visible, can be used as the calibration point. Our experimental results demonstrate that the gaze decomposition significantly improves estimation performance. Using only single gaze point calibration results in state-of-the-art performance for appearance-based eye trackers with calibration. Further improvement can be obtained by using multiple gaze points.  
	
	These two methods are based on the assumption that there exists person-dependent bias that can not be estimated from the images. It is known that there is a deviation between the visual axis and the optic axis of an eye, and that this deviation varies from person to person~\cite{atchison2000optics, guestrin2006general}. Our own experimental results confirm this finding. For a subject-independent estimator, the estimation bias varies significantly across subjects but seems to be relatively constant across different gaze angles for the same subject (see Fig.~\ref{fig:Error}). Thus, we decompose the gaze estimate (gaze of optic axis) into a subject-dependent bias term and the gaze of visual axis, which is subject-independent. The gaze of visual axis is estimated by a deep convolutional neural network. For calibration-free tracking, we set the bias to zero. For single gaze point calibration, we estimate the bias from images taken as the subject gazes at a single gaze point. 
	
	Our work is similar to~\cite{linden2018appearance} to some extent, where they included some subject-specific latent parameters in a deep network for calibration. Back propagation was applied for calibration. However, our network is specifically designed for single gaze point calibration, and the parameters to be calibrated (the bias) is easy to interpret and can be estimated by only a few images with little computational cost.
	
	We evaluate the performance of calibration-free subject-independent gaze estimation and the single gaze point calibration thoroughly. We evaluate through within- and cross-dataset settings on three public datasets: the MPIIGaze~\cite{zhang2015appearance}, the EYEDIAP~\cite{funes2014eyediap} and the ColumbiaGaze~\cite{smith2013gaze} datasets. Our experimental results demonstrate that as a calibration-free estimator, our proposed method outperforms the state-of-the-art methods that use single model on the MPIIGaze and the EYEDIAP datasets with a gain up tp $10.0\%$. Calibrating on only a few images (about $9$) can significantly reduce the estimation error, achieving a gain up to $35.6\%$ in comparison to the calibration-free estimator, and that the calibration is robust to the location of calibration points.

	
	\section{Related work}
	\subsection{Appearance-based gaze estimation}
	Methods for appearance-based gaze estimation directly regress from images to gaze estimate. In principal, given enough training data, they may be able to address the large variability in real-world situation, achieving relatively unconstrained gaze tracking.
	
	Past approaches to this problem have included k-Nearest Neighbors \cite{schneider2014manifold, sugano2014learning}, Support Vector Regression \cite{schneider2014manifold} and Random Forests \cite{sugano2014learning}. More recently, the application of deep CNNs to this problem has received increasing attention. Zhang \etal proposed the first deep CNN for gaze estimation in the wild \cite{zhang2015appearance,zhang2019mpiigaze}, which they showed improved accuracy significantly. To further improve the accuracy, others have proposed enhancements, such as employing the information outside the eye region \cite{krafka2016eye,zhang2017s}, focusing on the head-eye relationship \cite{deng2017monocular,ranjan2018light} and extracting better information from the eye images \cite{chen2018appearance,cheng2018appearance,lian2018multiview,park2018deep,yu2018deep}. 
	
	Krafka \etal proposed a CNN with multi-region input (an image of the face, images of both eyes and a face grid) to estimate the gaze target on screens of mobile devices \cite{krafka2016eye}. Zhang \etal proposed a network that takes the full face image as input and adopts a spatial weights method to emphasize features from particular face regions \cite{zhang2017s}. This work has shown that regions of the face other than the eyes also contain information about the gaze angle. 
	
	Some work has concentrated on the head-eye relationships. Deng and Zhu estimated the head pose in camera-centric coordinates, the gaze angles in head-centric coordinates, and then combined them geometrically~\cite{deng2017monocular}. Ranjan \etal applied a branching architecture, where parameters are switched according to a clustering of head pose angles into different groups~\cite{ranjan2018light}.
	
	Other work has focused on extracting better information from eye images. Cheng \etal studied the ``two eye asymmetry problem'': the estimation accuracy are different for both eyes. They proposed a novel network that relied on the high quality eye images for training~\cite{cheng2018appearance}. Yu \etal proposed to estimate the eye landmark locations and gaze directions jointly~\cite{yu2018deep}. Park \etal proposed to learn an intermediate pictorial representation of the eyes for better gaze estimation~\cite{park2018deep}. Chen and Shi proposed to use dilated-convolutions to extract features at high resolution, as large changes in gaze angle may result in only small changes in eye appearance~\cite{chen2018appearance}. Lian \etal improved accuracy by using images from multiple cameras~\cite{lian2018multiview}.
	
	\subsection{Personal calibration}
	\begin{figure*}
		\centering
		\subfloat[Mean (left) and SD (right) of yaw errors across 15 subjects.]{		
			\includegraphics[width=8.5cm,height=2.2cm]{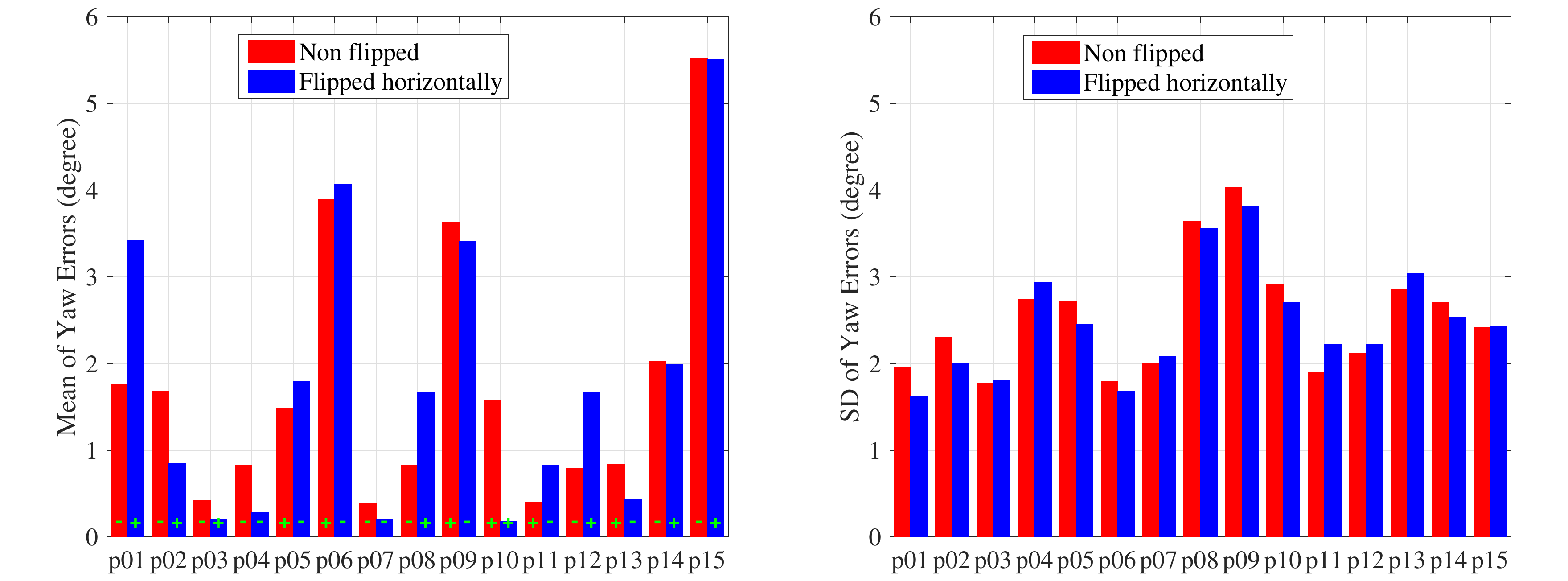}
			\label{fig:1-a}	
		}
		\hspace{-0.1cm}
		\subfloat[Mean (left) and SD (right) of pitch errors across 15 subjects.]{		
			\includegraphics[width=8.5cm,height=2.2cm]{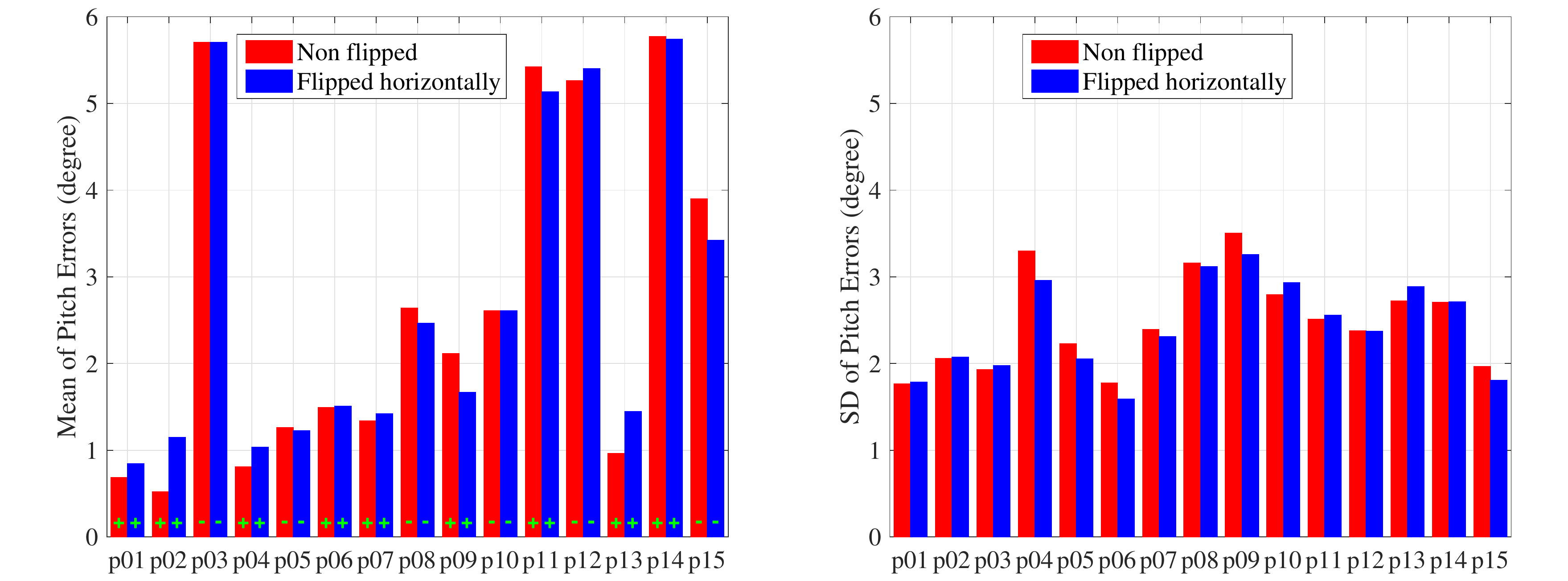}
			\label{fig:1-b}
		}
		
		\subfloat[Estimated angles versus ground truth angles of p06.]{		
			\includegraphics[width=8.5cm,height=2.2cm]{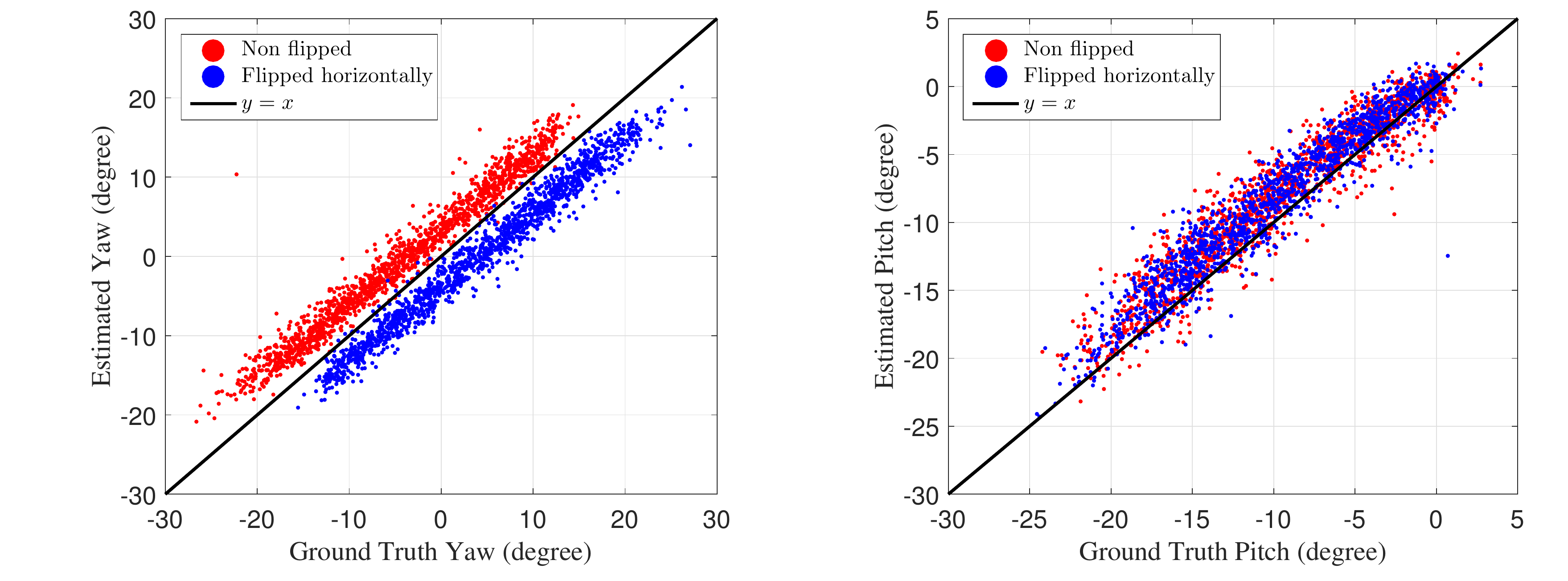}
			\label{fig:1-c}	
		}
		\hspace{-0.1cm}
		\subfloat[Estimated angles versus ground truth angles of p12.]{		
			\includegraphics[width=8.5cm,height=2.2cm]{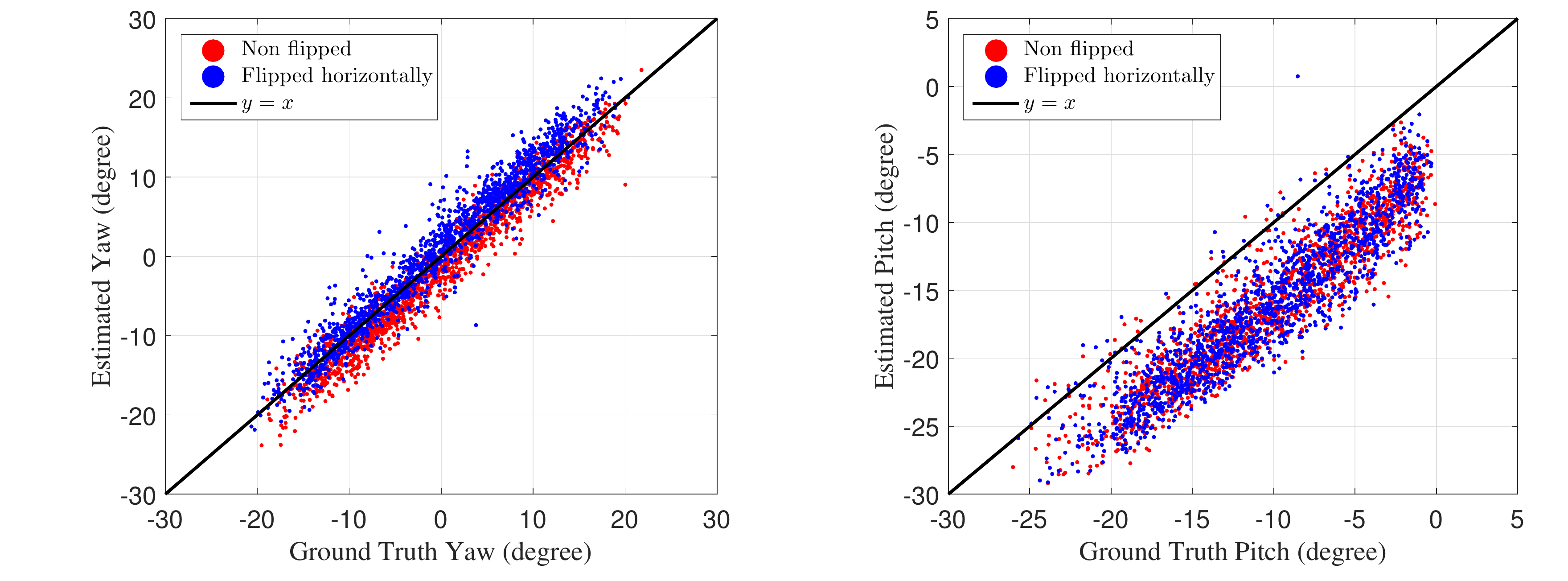}
			\label{fig:1-d}
		}
		\caption{Error analysis of a subject-independent estimator on the MPIIGaze dataset. Mean and standard deviation of the yaw (a) and pitch (b) angles for each subject. The green $+$ or $-$ indicate the sign of the mean. Scatter plots of the estimated angles versus the ground truth of two subjects, p06 (c) and p12 (d). Half of the images are non-flipped (red) and half are flipped horizontally (blue). There exists person-dependent bias that vary across subjects, but is quite constant across gaze angles. Better viewed in color and zoomed in.}
		\label{fig:Error}
	\end{figure*}
	Calibration for appearance-based methods has received little attention to date. To our knowledge, the first implementation of personal calibration was in iTracker~\cite{krafka2016eye}. After training a subject independent network, they calibrated by training a person-dependent Support Vector Regressor on the features from the last fully-connected layer using images collected as the subject gazed at 13 different locations. Error was reduced by about $20\%$ when calibrated on the full set of 13 points, but increased when only calibrated on a subset of 4 points, most likely due to overfitting. Lind{\'e}n \etal proposed a network that includes some person-dependent latent parameters~\cite{linden2018appearance}. During training, these parameters are learned from the training data. During testing, these parameters can be estimated by minimizing the estimation error in the calibration set with the rest of the network fixed. The calibration set required ranged from 45-100 images and was collected as the subject gazed at multiple points. Liu \etal proposed a differential approach for calibration \cite{liu2018differential}. They trained a subject-independent Siamese network to estimate the gaze angle difference between two images of the same subject. During testing, they used this network to estimate gaze differences between the input image and nine calibration images taken as the subject gazed at different locations, and averaged the resulting estimates. 
	
	In comparison to the work described above, our proposed method achieves better accuracy, while requiring a smaller number of images taken as the subject gazes at only a single point.
	
	\section{Methodology}
	\subsection{Analysis of estimation error}
	Fig.~\ref{fig:Error} analyzes estimation error on the MPIIGaze~\cite{zhang2015appearance} dataset made by a state-of-the-art subject-independent gaze estimator~\cite{chen2018appearance}. Fig.~\ref{fig:Error}(a) and (b) show the mean and standard deviation (SD) of the yaw and pitch error for different subjects and for both original and horizontally flipped images. These results show that the errors in both yaw and pitch angles are generally biased. The bias vary across subjects, whereas the SDs are relatively stable. When the images are flipped horizontally, the yaw bias typically has similar magnitude,but different sign, while the pitch bias remains similar. The mean squared bias across subjects (16.2 $\mathrm{deg}^2$) exceeds the mean intra-subject variance (12.9 $\mathrm{deg}^2$), indicating that the bias is a significant contributor to the error.
	
	Fig.~\ref{fig:Error}(c) and (d) compare the estimated angles and the ground truth of two subjects. The scatter plots indicate that the bias is quite constant across gaze angles, but varies between subjects. As the data of each subject  exhibits considerable variability in illumination, we hypothesize that a major determinant of this bias is due to user differences.
	\begin{figure*}
		\centering
		\includegraphics[width=15.5cm,height=4.9cm]{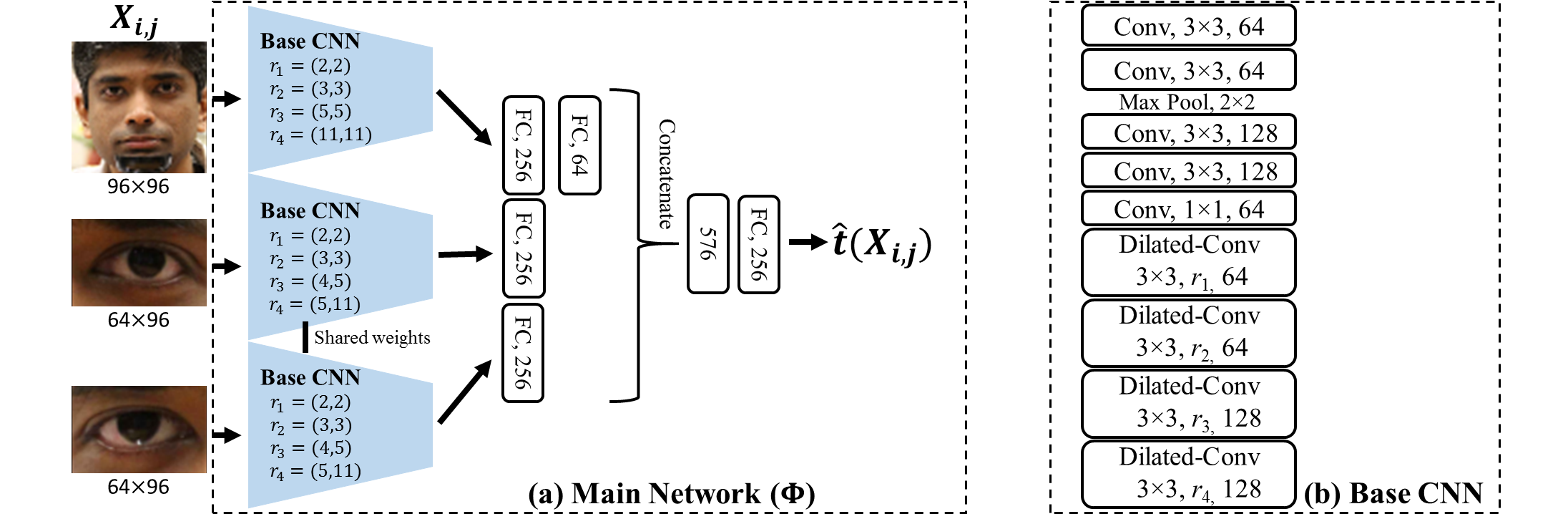}
		\caption{Architecture of the proposed network. (a) The main network that outputs $\hat{t}(X_{i,j})$ based on the input image $X_{i,j}$. (b) The base CNN is the basic component of (a). FC denotes fully-connected layers, Conv denotes convolutional layers, Dilated-Conv denotes dilated-convolutional layers, and $r$ is the dilation rate of the dilated-convolutional layer.}
		\label{fig:Archi}
	\end{figure*}
	\subsection{Gaze decomposition}
	Motivated by the previous findings, we assume that there exists person-dependent bias that can not be estimated from the images. This assumption is supported by the fact that there exists a person-dependent deviation between the visual axis (the line connecting the nodal point with the fovea) and the optic axis (the line connecting the nodal point with the pupil center), but is not observable from the appearance of the eye~\cite{atchison2000optics,guestrin2006general}. This assumption can be formally formulated as follows, where we use $i$ to denote the index of subject and $j$ the index of image:
	\begin{equation}
	X_{i,j}=\mathcal{F}(c_i, t_{i,j}),
	\label{gazeModel1}
	\end{equation}
	\begin{equation}
	g_{i,j}=t_{i,j}+b_i,
	\label{gazeModel2}
	\end{equation}
	where $X_{i,j}$ an image of subject $i$ with index $j$, $\mathcal{F}$ an arbitrary function, $c_i$ the appearance feature of subject $i$, $t_{i,j}$ the gaze of visual axis, $g_{i,j}$ the true gaze (gaze of optic axis) expressed as yaw and pitch, and $b_i$ the person-dependent bias. One can not estimate $b_i$ from $X_{i,j}$.
	
	Under this assumption, we decompose gaze estimates, $\hat{g}$, into the sum of a subject-dependent bias term, $\hat{b}$, and a subject-independent gaze of visual axis, $\hat{t}$:
	\begin{equation}
	\hat{g}(X_{i,j})=\hat{t}(X_{i,j})+\hat{b}_i.
	\label{dec}
	\end{equation}
	Note that a model which directly estimates the gaze angles is equivalent to a model with $\hat{b}_i \equiv 0, \forall i$. We describe below how to estimate $\hat{b}_i$ under different scenarios: training, calibration-free estimation and calibration.
	
	\subsection{The proposed network}
	The architecture of our proposed network is presented in Fig.~\ref{fig:Archi}. The general architecture is inspired by iTracker \cite{krafka2016eye} and Dilated-Net \cite{chen2018appearance}. It takes an image of the face and images of both eyes as input. The output of the network $\hat{t}(X_{i,j})$ is used to generate gaze estimates according to \eqref{dec}.
	
	The input images $X_{i,j}$ are first fed to three base CNNs. The architecture of the base CNN is shown in Fig.~\ref{fig:Archi}(b). It has five convolutional layers, one max-pooling layer and four dilated-convolutional layers \cite{yu2015multi} with different dilation rate, $r$. The strides for all (dilated-) convolutional layers are $1$. The weights of the first four convolutional layers are transfered from VGG-16 \cite{simonyan2014very} pre-trained on the ImageNet dataset \cite{deng2009imagenet}. The two base CNNs that take the eyes as input share the same weights. The feature maps extracted by the base CNNs are then fed to fully-connected (FC) layers, concatenated, fed to another FC layer followed by a linear output layer to output $\hat{t}(X_{i,j})$. We denote the parameters of this network as $\Phi$.
	
	Rectified Linear Units (ReLUs) are used as the activation function. Zero-padding is applied to regular convolutional layers and no padding is applied to dilated-convolutional layers. Batch renormalization layers \cite{ioffe2017batch} are applied to all layers trained from scratch. Dropout layers with dropout rates of $0.5$ are applied to all FC layers.
	\\
	\\
	\noindent\textbf{Training.}	Based on~\eqref{gazeModel1}\eqref{gazeModel2}\eqref{dec}, we train the network by solving the following optimization problem:
	\begin{equation}
	\min_{\Phi,\beta} \left(\sum_{i,j}\begin{Vmatrix}
	g_{i,j}-\hat{t}(X_{i,j};\Phi)-\beta_i
	\end{Vmatrix}_2^2
	+ \lambda | \sum_i \beta_i | \right)
	\label{train_t}
	\end{equation}
	where $\beta_i$ are estimates of the bias for subjects in the training set  which are learned. The second term is a regularizer that ensures average subject-dependent bias over the training set is zero. Since this mean bias is arbitrary, the training is insensitive to the value of $\lambda$. The $\beta_i$ could also be estimated by calculating the individual mean over the training set, but this would be time consuming, given the large size of training set, especially since we use online data augmentation. 
	
	We use Adam optimizer with default parameters in TensorFlow and a batch size of $64$. An initial learning rate of $0.001$ is used. It is divided by $10$ after every ten epochs. The training proceeds for $35$ epochs. We apply online data augmentation including random cropping, scaling, rotation and horizontal flipping. As the bias changes if the images are flipped horizontally, we considered the non-flipped and flipped images as belonging to different subjects.
	\\
	\\
	\noindent\textbf{Testing and calibration.}
	During testing, gaze estimates are generated according to~\eqref{dec}. For a new subject~$m$, if no calibration images are available, we set $\hat{b}_m = 0$. 
	
	With calibration, there is a trade-off between the complexity of the calibration dataset ($\mathcal{D}_c$) and accuracy. Simpler $\mathcal{D}_c$ are desirable, as they are easier to collect, but result in lower accuracy, since the statistics of the calibration set might be significantly different from that of the test set. Some existing methods, e.g. linear adaptation~\cite{liu2018differential} and fine-tuning, may yield bad performance as they inexplicitly assume that the statistics of the calibration set is similar to that of the test set. The challenge we address is achieving the best accuracy for the lowest complexity. We focus on two measures of calibration set complexity: the variability in gaze locations, e.g. single gaze point calibration (\textbf{SGPC}) versus multiple gaze point calibration (\textbf{MGPC}), and the number of images. 
	
	Given a calibration set $\mathcal{D}_c$, which includes image-gaze pairs for a subject~$m$, $\{(X_{m,j}, g_{m,j}),j=1,2,\dots,|\mathcal{D}_c|\}$, we set $\hat{b}_m = \tilde{b}_m$, where 
	\begin{equation}
	{\tilde{b}}_m=\frac{1}{|\mathcal{D}_c|}\sum_{X_{m,j}\in \mathcal{D}_c} (g_{m,j}-\hat{t}(X_{m,j})), 
	\label{cali1}
	\end{equation}
	where $|\cdot|$ represents cardinality. We only use two parameters for calibration to avoid over-fitting. Unlike~\cite{linden2018appearance}, which requires back propagation for calibration, our proposed method only requires forward propagation, reducing computational cost.
	\\
	\\
	\noindent\textbf{Preprocessing.} We apply the modified data normalization method introduced in \cite{zhang2018revisiting}. This method rotates and scales an image so that the resulting image is taken by a virtual camera facing towards a reference point on the face at a fixed distance and canceling out the roll angle of the head. The images are normalized by perspective warping, converted to gray scale and histogram-equalized. For automatically detected landmarks we use dilb \cite{king2009dlib}.
	\section{Experiments}
	We evaluated our proposed network through cross-subject evaluation in both within- and cross-dataset settings. We evaluated the calibration-free estimation error and the performance of calibration.
	
	\subsection{Datasets}
	We used the MPIIGaze~\cite{zhang2015appearance} and the EYEDIAP~\cite{funes2014eyediap} datasets for within-dataset evaluation. We trained on the MPIIGaze and tested on the ColumbiaGaze~\cite{smith2013gaze} for cross-dataset evaluation.
	
	The MPIIGaze contains full face images of 15 subjects (six female, five with glasses). We trained and tested on the ``Evaluation Subset'', which contains $3,000$ randomly selected images for each subject. Within this subset, half of the images are flipped horizontally. The reference point for image normalization was set to the center of the face.
	
	The EYEDIAP contains full face videos with three gaze targets (continuous screen target, discrete screen target and floating target) and two types of head pose (static and dynamic). We used the data from screen targets, which involve 14 subjects (three female, none with glasses). The reference point for image normalization was set to the midpoint of both eyes.
	
	The ColumbiaGaze has $5,800$ full face images of $56$ subjects ($24$ female, $21$ with glasses). For each subject, images were collected for each combination of five horizontal head poses $(0^\circ, \pm15^\circ, \pm30^\circ)$, seven horizontal gaze directions $(0^\circ, \pm5^\circ, \pm10^\circ, \pm15^\circ)$ and three vertical gaze directions $(0^\circ, \pm10^\circ)$. We excluded the images with $10^\circ$ vertical gaze directions from the ColumbiaGaze dataset for these evaluations, since the MPIIGaze dataset mainly covers pitch angles from $-20^\circ$ to $0^\circ$. For SGPC, we used the five images with the same gaze target, but with different head pose.
	
	\subsection{Within-dataset evaluation}
	\begin{figure}
		\centering
		\includegraphics[width=9cm]{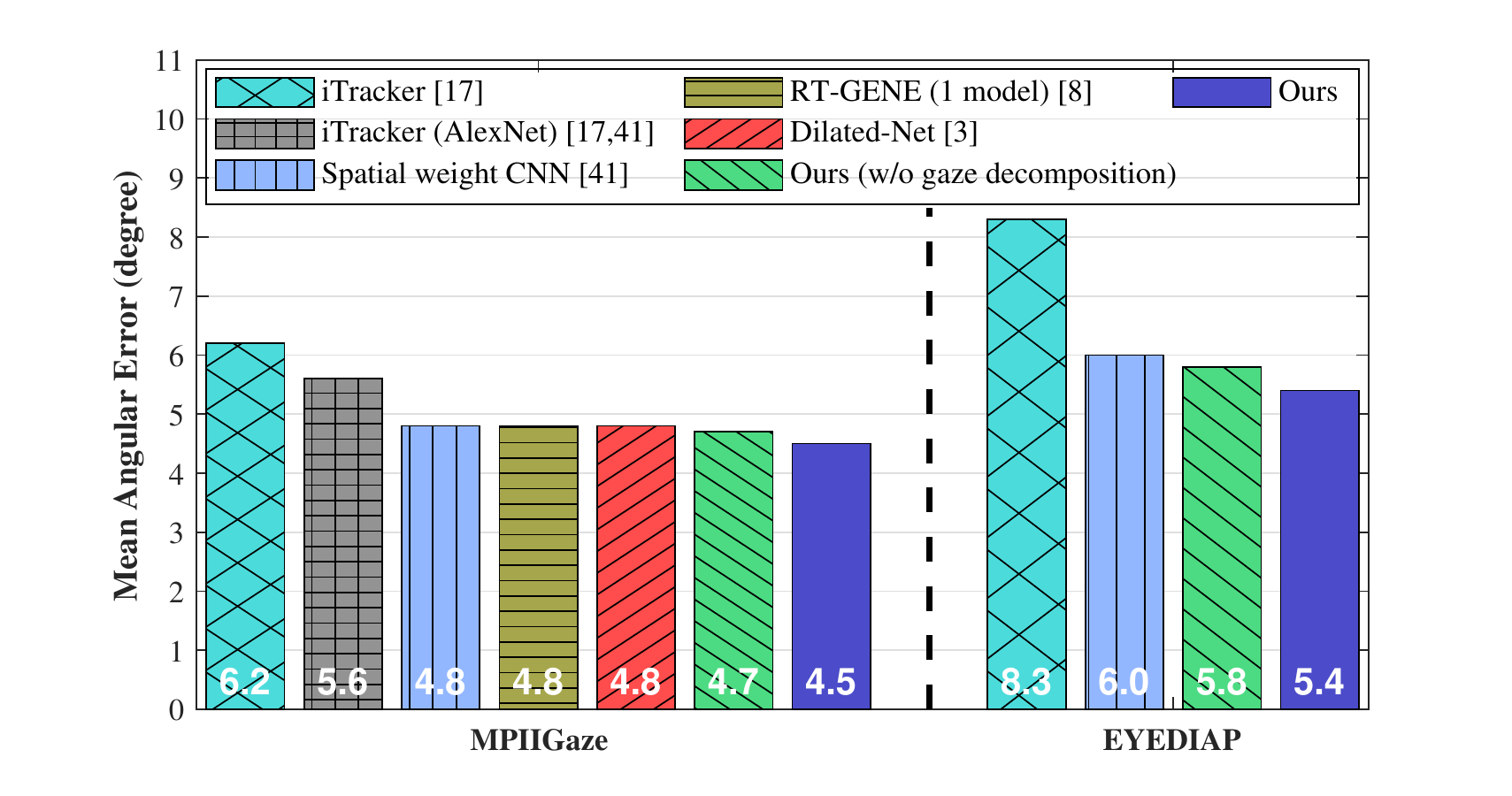}
		\caption{Mean angular error of calibration-free estimation on the MPIIGaze and the EYEDIAP dataset.}
		\label{fig:CalFree}
	\end{figure}
	
	\noindent\textbf{Calibration-free estimation.} For MPIIGaze, we conducted 15-fold leave-one-subject-out cross-validation as~\cite{zhang2017s,chen2018appearance}. For EYEDIAP,  we followed the protocol described in~\cite{zhang2017s}, i.e., five-fold cross-validation on four VGA videos (both screen targets with both types of head pose) sampled at 2 fps (about 1,200 images per subject).
	
	We compared with several baselines: iTracker~\cite{krafka2016eye,zhang2017s}, spatial weights CNN~\cite{zhang2017s}, RT-GENE~\cite{fischer2018rt} and Dilated-Net~\cite{chen2018appearance}. All of these methods used face images (or face plus eye images) as input. We also trained a network without gaze decomposition for ablation study.
	
	The results are shown in Fig.~\ref{fig:CalFree}. On the MPIIGaze dataset, our proposed network achieved $4.5^\circ$ mean angular error , which outperformed the state-of-the-art $4.8^\circ$~\cite{chen2018appearance,fischer2018rt,zhang2017s} by $6.3\%$. On the EYEDIAP dataset, it achieved $5.4^\circ$, which outperformed the state-of-the-art $6.0^\circ$~\cite{zhang2017s} by $10.0\%$. The gain may partly due to the use of the modified data normalization method~\cite{zhang2018revisiting} and our proposed gaze decomposition. Note that for consistency with the other models, we only report the accuracy for RT-GENE without ensembling. Using an ensemble of four models, RT-GENE achieved an accuracy of $4.3^\circ$ on MPIIGaze. Applying ensembling to our model did not yield better performance.
	\\
	\\
	\noindent\textbf{Calibration.} To perform single gaze point calibraion (SGPC), we first randomly select a calibration point in the 2D (yaw, pitch) gaze space. We then randomly selected $|\mathcal{D}_c|$ images from the test set whose true gaze angles differed from the calibration point by less than $2^\circ$ as $\mathcal{D}_c$. We calibrated on $\mathcal{D}_c$ according to~\eqref{cali1}, where we set the true gaze angle to the mean of the gaze angles in the calibration set. We tested on the images not belonging to $\mathcal{D}_c$. We discarded a calibration point if less than $|\mathcal{D}_c|$ images met the $2^\circ$ requirement. Note that one gaze point can have multiple calibration samples.
	
	For	multiple gaze point calibration (MGPC), we randomly selected $|\mathcal{D}_c|$ images from the test set as $\mathcal{D}_c$, calibrated according to~\eqref{cali1} and tested on the images not belonging to $\mathcal{D}_c$.
	
	We also calculated the lower bound performance of our proposed method, termed $\underline{E}$, by estimating the bias from all images in the test set and evaluating on the test set. We denote the mean angular error of the calibration-free subject-independent tracking as $\overline{E}$.
	\\
	\\
	\noindent\textbf{Calibration results of MPIIGaze.}
	\begin{figure}
		\centering
		\includegraphics[width=9cm]{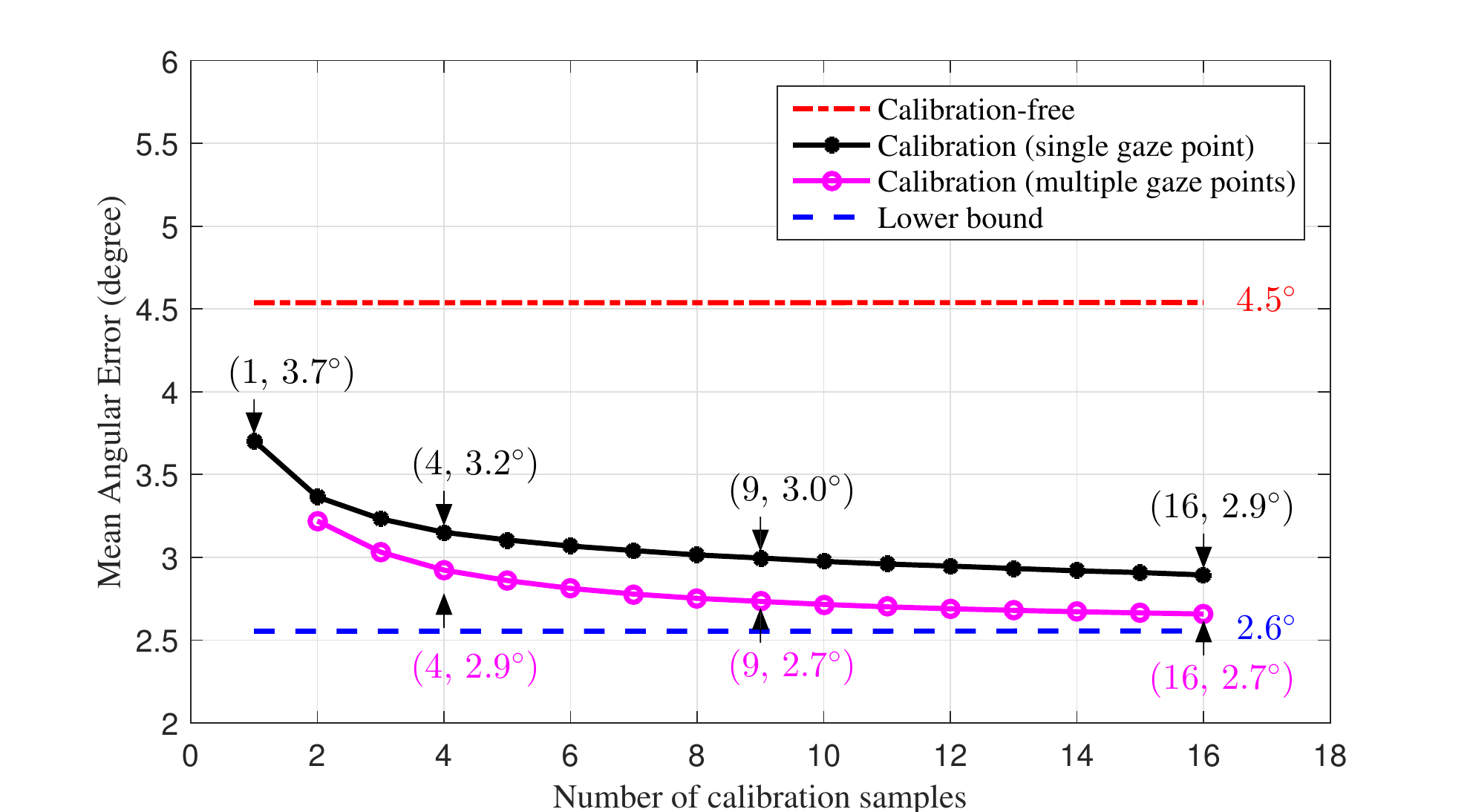}
		\caption{Mean angular error as a function of number of calibration samples on the MPIIGaze dataset.}
		\label{fig:MPII1}
	\end{figure}
	In the MPIIGaze, since half of the images were flipped and we considered the data from the non-flipped and flipped images separately when computing the bias and the accuracy, each subject generates two accuracies (30 in total). Note that none of the flipped/non-flipped images of the test subject were included in the training set. To evaluate the overall performance of SGPC and MGPC across different calibration points and subjects, we report average errors computed over $5,000$ trials. We also varied the size of $\mathcal{D}_c$. 
	
	\begin{table}[!t]
		\begin{center}
			\begin{tabular}{c|c|cccccc}
				\cline{1-8}
				&$|\mathcal{D}_c|$&1&4&8&16&64&128\\
				\cline{1-8}
				\multicolumn{2}{c|}{\small{Cali.-free (Ours)}}&\multicolumn{6}{c}{4.5}\\
				\cline{1-8}
				\multirow{6}*{SGPC}&FC&4.8&6.9&7.9&8.4&&\\
				\cline{2-8}
				&LA~\cite{liu2018differential}&\textit{NA}&15.1&14.9&14.3&&\\
				\cline{2-8}
				&LP~\cite{linden2018appearance}&4.2&3.9&3.8&3.7&&\\
				\cline{2-8}
				&DF~\cite{liu2018differential}&4.2&3.6&3.4&3.3&&\\
				\cline{2-8}
				&\tabincell{c}{Ours\\(w/o GD)}&4.2&3.6&3.5&3.4&&\\
				\cline{2-8}
				&Ours&\textbf{3.7}&\textbf{3.2}&\textbf{3.0}&\textbf{2.9}&&\\
				\cline{1-8}
				\multirow{6}*{MGPC}&FC&&5.8&4.6&3.5&2.7&2.5\\
				\cline{2-8}
				&LA~\cite{liu2018differential}&&5.2&3.2&2.8&\textbf{2.5}&\textbf{2.4}\\
				\cline{2-8}
				&LP~\cite{linden2018appearance}&&3.4&3.2&3.0&2.9&2.9\\
				\cline{2-8}
				&DF~\cite{liu2018differential}&&3.3&3.1&3.0&2.6&2.6\\
				\cline{2-8}
				&\tabincell{c}{Ours\\(w/o GD)}&&3.3&3.1&3.0&2.9&2.9\\
				\cline{2-8}
				&Ours&&\textbf{2.9}&\textbf{2.8}&\textbf{2.7}&2.6&2.6\\
				\cline{1-8}
			\end{tabular}
		\end{center}
		\caption{Estimation Error ($^\circ$) of Different Calibration Methods on MPIIGaze. GD means gaze decomposition.}
		\label{tab:MPII}
	\end{table}
	Fig.~\ref{fig:MPII1} presents the mean angular error. For both methods, as the complexity of calibration set increases (number of calibration samples or the number off images increases), the error decreases. We only show the results calibrated on no more than $16$ samples because the dataset does not has enough samples. SGPC reduced the estimation error significantly. For example, when calibrated on $16$ samples, it reduced the error by $1.6^\circ$ ($35.6\%$) in comparison to calibration-free estimation. MGPC led to further reductions. On average, the gap between SGPC to MGPC was about $0.2^\circ$ ($4.4\%$ of the error of the calibration-free estimator). This gap suggests that the bias is not constant across different gaze angles. 
	
	We compared our approach with several existing calibration method: fine-tuning the last FC layer (\textbf{FC}), linear adaptation (\textbf{LA})~\cite{liu2018differential}, the differential method (\textbf{DF})~\cite{liu2018differential} and fine-tuning the latent parameters (\textbf{LP})~\cite{linden2018appearance}. FC and LA were directly applied to our network with gaze decomposition. DF and LP were re-implemented using our architecture. These results are comparable as they are all applied to the same backbone network. 
	
	TABLE~\ref{tab:MPII} shows that our proposed method performs the best for low complexity calibration sets. It outperforms other methods for all tested cases of SGPC and for MGPC when the number of images is less than or equal to 16. For example, for SPGC with 16 samples, our proposed algorithm reduced the error by $12.1\%$ when compared to the second best method (DF). We attribute this better performance to the small number of parameters, which avoids overfitting. Unlike LP, our method acts directly on the gaze estimates, making it more effective if the primary cause of error is a subject-dependent bias. As the complexity of the calibration set increases, other methods (e.g. LA) will outperform ours. However, we believe that obtaining more than 64 images for calibration is very time-consuming in most real-world applications.
	\\
	\\
	\noindent\textbf{Robustness to calibration location}.
	\begin{figure}
		\centering
		\includegraphics[width=9cm]{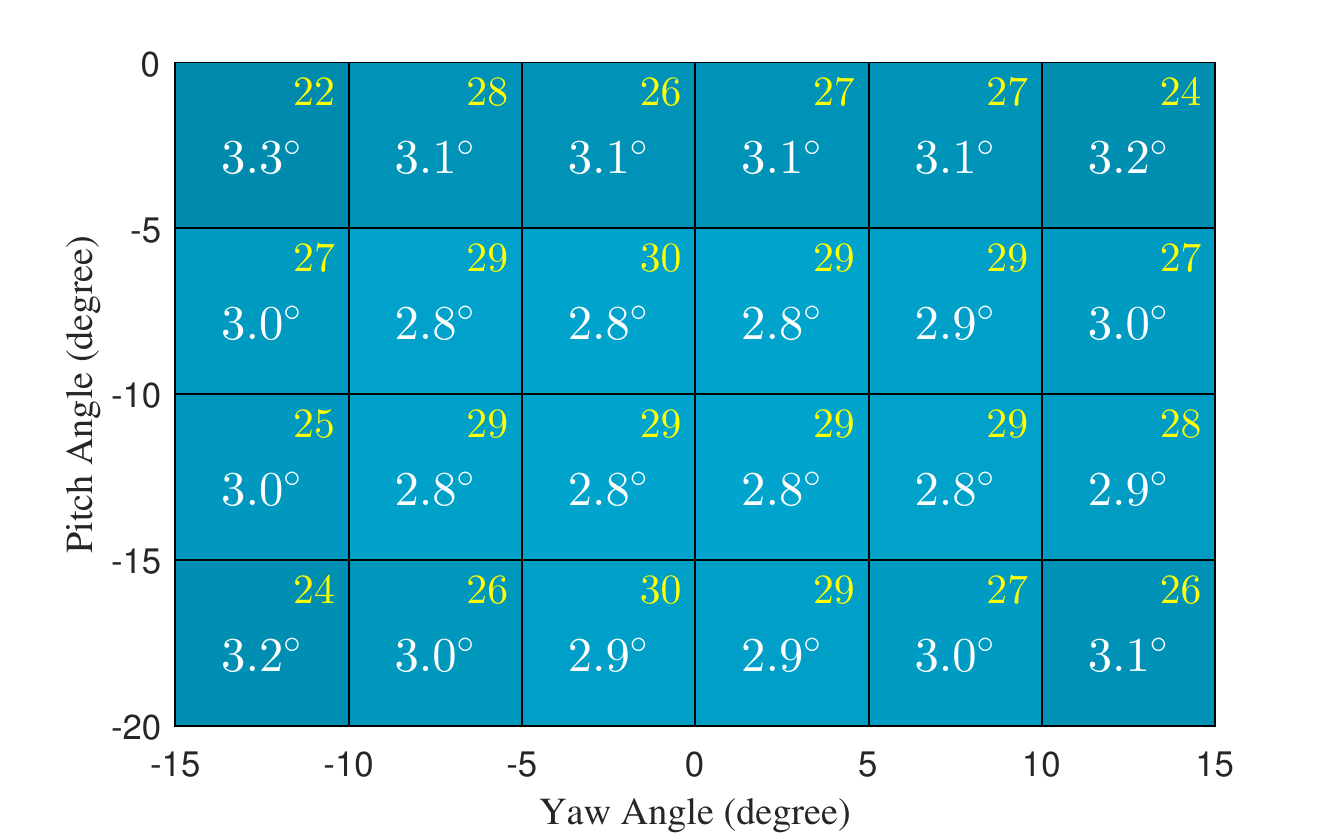}
		\caption{Mean angular error (white) and the number of subjects that benefit among $30$ subjects (yellow) when calibrated at different locations on the MPIIGaze. Each $5^\circ \times 5^\circ$ region includes a $10 \times 10$ grid of calibration points. $\overline{E}=4.5^\circ, \underline{E}=2.6^\circ, |\mathcal{D}_c|=9$.}
		\label{fig:MPII2}
	\end{figure}
	To evaluate the robustness of SGPC to the location of calibration point, we create a grid of calibration points which were uniformly distributed in the (yaw, pitch) space with a step of $0.5^\circ$ in each axis. We calibrated at each point in the grid, and tested on the remaining testing images. We set $|\mathcal{D}_c|=9$, since the error reduction from our above experiment begins to saturate at this point. 
	
	Fig.~\ref{fig:MPII2} presents the mean angular error for calibration points located in different $5^\circ \times 5^\circ$ regions. We only show the results of this region because there are not enough data outside this region. The error achieved by calibrating at the center of the gaze range is lower than the error achieved by calibrating at the boundary. However, the standard deviation over locations is only $0.15^\circ$, indicating that SGPC is quite robust to the location of calibration point. At least $22$ out of $30$ subjects ($73.3\%$) benefited from it. 
	\\
	\\
	\noindent\textbf{Calibrtion results of EYEDIAP}. We conducted leave-one-subject-out cross-validation on two VGA videos (continuous screen target with both types of head pose) sampled at 15 fps (about $3,500$ images per subject). We chose these data because they had sufficient annotation to remove outliers, e.g. images during blinking. 
	
\begin{figure}
		\centering
		\includegraphics[width=9cm]{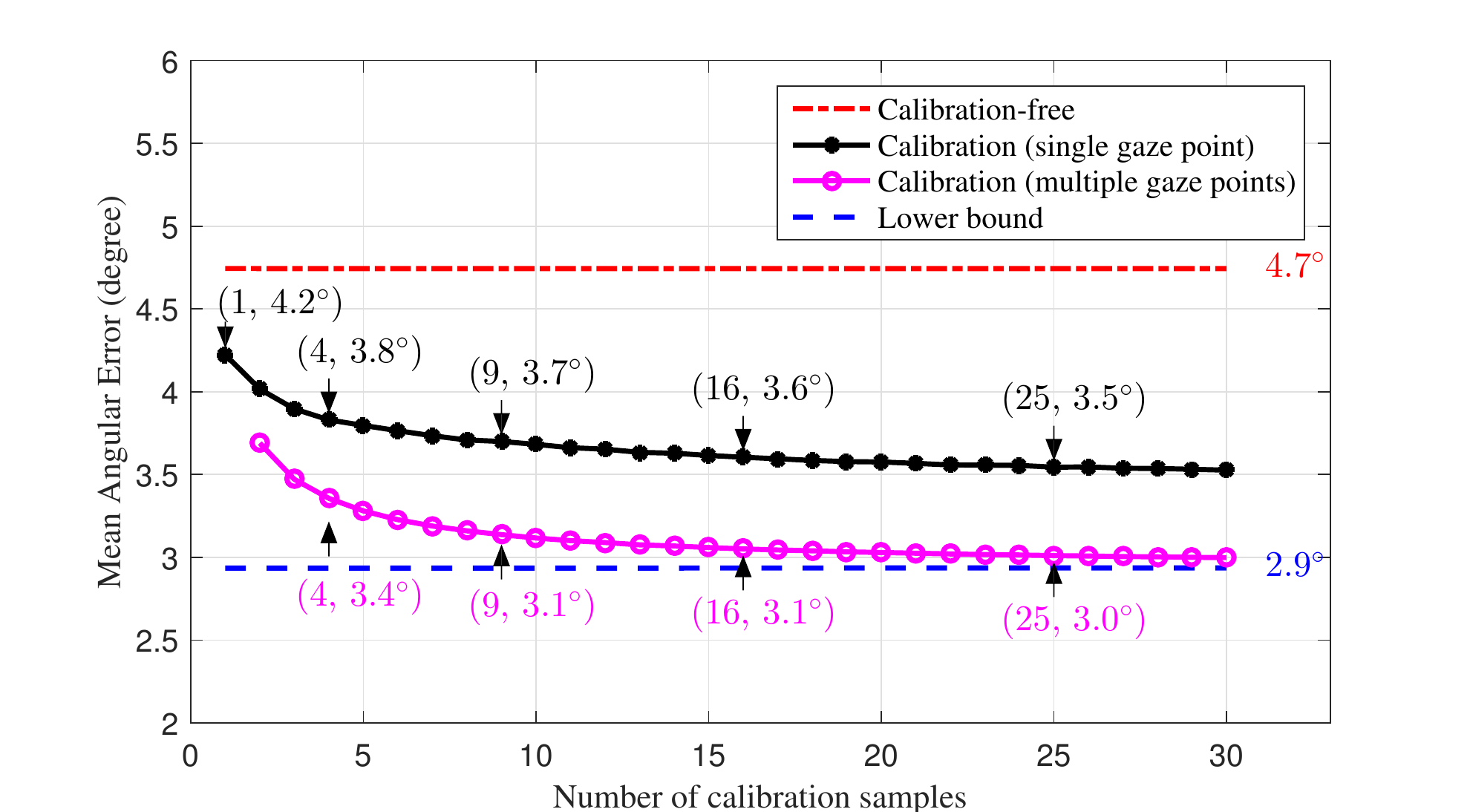}
		\caption{Mean angular error after calibration as a function of number of calibration samples on the EYEDIAP dataset.}
		\label{fig:EYEDIAP1}
	\end{figure}

\begin{figure}
		\centering
		\includegraphics[width=9cm]{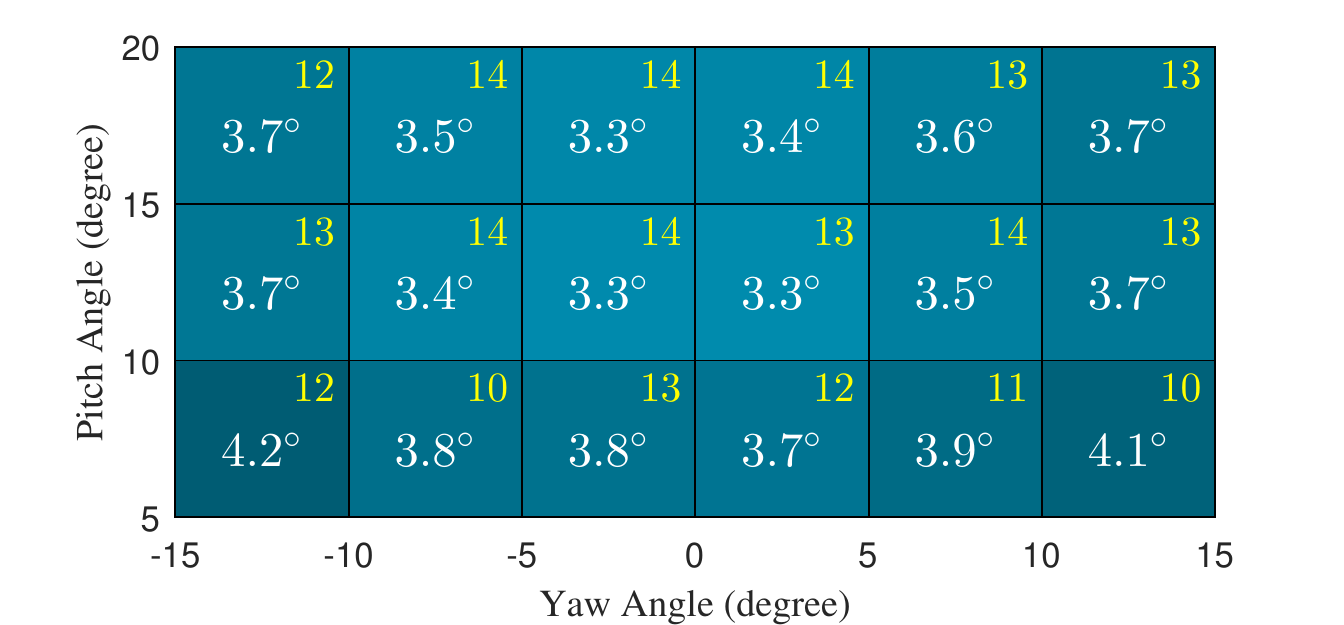}
		\caption{Mean angular error (white) and the number of subjects that benefit among $14$ subjects (yellow) when calibrated at different locations on the EYEDIAP dataset. $\overline{E}=4.7^\circ, \underline{E}=2.9^\circ, |\mathcal{D}_c|=9$.}
		\label{fig:EYEDIAP2}
	\end{figure}

	The results are shown in Fig.~\ref{fig:EYEDIAP1} and Fig.~\ref{fig:EYEDIAP2}. The findings are generally consistent to the ones of the MPIIGaze. The overall performance is worse than that on the MPIIGaze, most likely due to the lower resolution and higher variability of head pose. In Fig.~\ref{fig:EYEDIAP1}, SGPC reduced the error significantly. For example, a $1.2^\circ$ ($25.5\%$) improvement was achieved when calibrated on $25$ samples. In Fig.~\ref{fig:EYEDIAP2}, $0.5^\circ$ ($10.6\%$) to $1.4^\circ$ ($29.8\%$) improvement was achieved across different locations. At least $10$ out of $14$ subjects ($71.4\%$) benefited. The SD over locations was $0.26^\circ$.
	
	As a calibration-free estimator, the mean error achieved in this experiment ($4.7^\circ$) was better than that in the previous experiment ($5.4^\circ$, Fig.~\ref{fig:CalFree}), most likely due to that there were much more training samples in this experiment as it ran at 15 fps instead of 2 fps, and that outliers were removed in this experiment given the extra annotation on this subset.
	
	\subsection{Cross-dataset evaluation}
	We trained on the MPIIGaze and tested on the ColumbiaGaze. Fig.~\ref{fig:Colum} shows the calibration performance for different calibration gaze targets, where $\overline{E}=5.5^\circ$ and $\underline{E}=4.1^\circ$, indicating the difficulty of cross-dataset evaluation. Consistent with our previous results, the performance was poor when calibrated at the four gaze targets at the boundary, i.e., samples with horizontal directions $\pm15^\circ$. However, for the middle ten calibration points, SGPC reduced the error by $0.8^\circ$ ($14.5\%$) to $1.1^\circ$ ($20.0\%$) in comparison to the calibration-free estimator. Between $46$ to $51$ out of $56$ subjects ($82.1\%$ to $91.1\%$) benefited.
	
	Our method also outperformed LP by $0.8^\circ$ ($14.5\%$) and DF by $0.6^\circ$ ($11.3\%$) in this cross-dataset evaluation (see Table~\ref{tab:ap_error} in Appendix).
	
	\subsection{Ablation study of gaze decomposition}
	We trained a network without gaze decomposition by setting $\beta_i\equiv0,\forall i$ in \eqref{train_t} during training. We refer to it as \textbf{ND} and the proposed network with gaze decomposition as \textbf{D}. In Fig.~\ref{fig:CalFree}, D network outperformed ND network in calibration-free estimation on the MPIIGaze and EYEDIAP datasets.
	
\begin{figure}
		\centering
		\includegraphics[width=9cm]{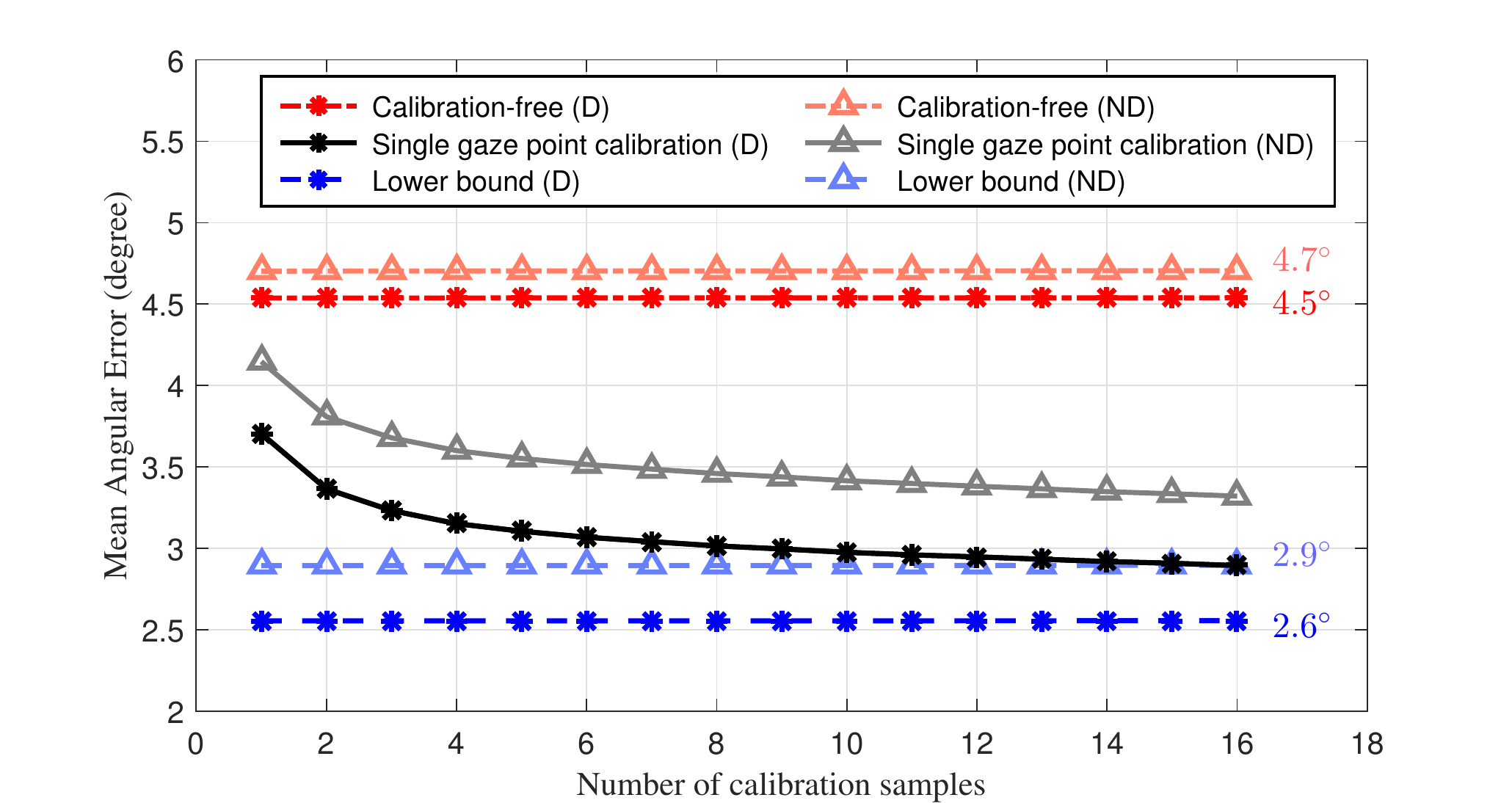}
		\caption{Comparison of the networks with/without gaze decomposition (D and ND) on the MPIIGaze dataset.}
		\label{fig:AblaMPII}
	\end{figure}

	We evaluated both networks on the MPIIGaze by leave-one-subject-out cross-validation. Fig.~\ref{fig:AblaMPII} presents the results over calibration points and subjects. The D network achieved lower errors in all metric, i.e., calibration-free, SGPC and calibration lower bound. For SGPC, on average, D network achieved a gain about $0.35^\circ$ from the ND network. The D network achieved smaller calibration lower bound ($0.3^\circ$), indicating that the intra-subject variance was reduced by gaze decomposition. These results are consistent to our model~\eqref{gazeModel1}\eqref{gazeModel2}: as the person-dependent bias can not be estimated from the images, removing the bias during training according to \eqref{train_t} makes $\hat{t}$ generalizes better.
	\begin{figure}
		\centering
		\includegraphics[width=9cm]{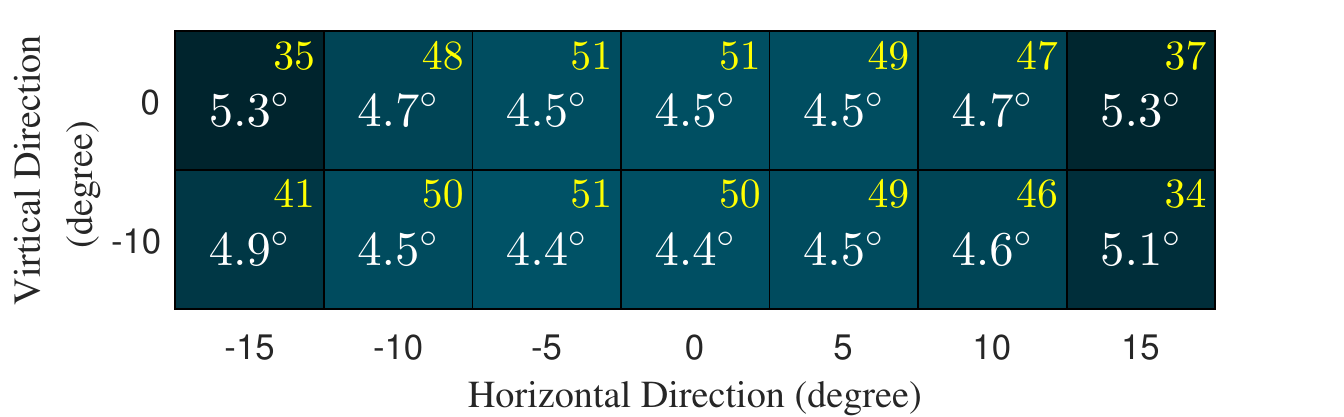}
		\caption{Mean angular error (white) and the number of subjects that benefit among $56$ subjects (yellow) when calibrated at different gaze targets. Trained on the MPIIGaze and test on the Columbia dataset. $\overline{E}=5.5^\circ, \underline{E}=4.1^\circ, |\mathcal{D}_c|=5$.}
		\label{fig:Colum}
	\end{figure}
	
\begin{figure}
		\centering
		\includegraphics[width=9cm]{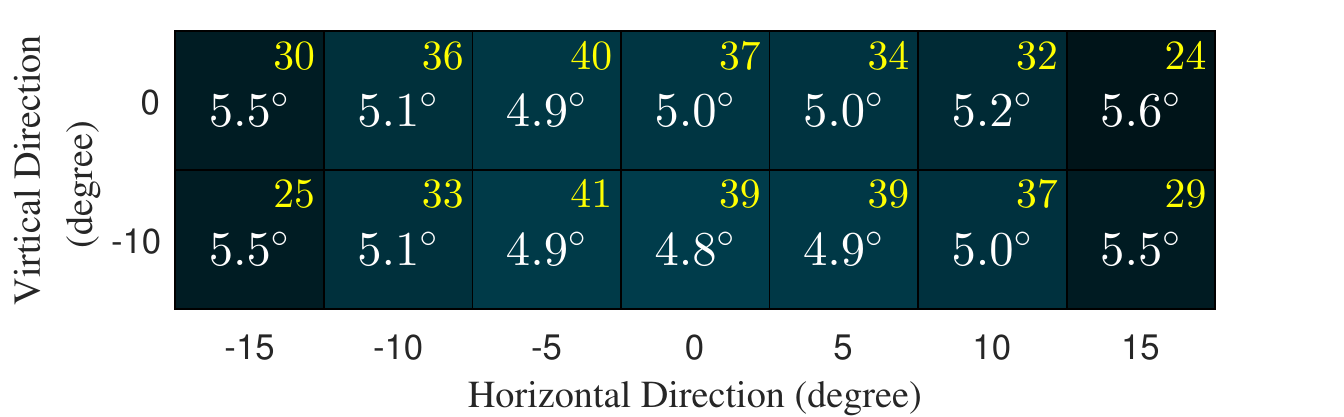}
		\caption{Mean angular error achieved by the network without gaze decomposition when calibrated at different gaze targets. Trained on the MPIIGaze and test on the Columbia dataset. $\overline{E}=5.5^\circ, \underline{E}=4.5^\circ, |\mathcal{D}_c|=5$.}
		\label{fig:AblaColum}
	\end{figure}

	We also repeated the cross-dataset evaluation for the ND network. The results are shown in Fig.~\ref{fig:AblaColum}, where $\overline{E}=5.5^\circ$ and $\underline{E}=4.5^\circ$. Comparing with the results of the D network in Fig.~\ref{fig:Colum}, both networks achieved the same $5.5^\circ$ in calibration-free tracking. However, on average, a gain about $0.4^\circ$ was achieved by gaze decomposition when both networks were calibrated on five samples.
	\subsection{Consistency of the learned bias}
	We evaluated whether the learned biases, i.e., $\beta_i$ in~\eqref{train_t} were consistent for the same subject using leave-one-subject-out (15 fold) cross-validation on the MPIIGaze dataset. For each subject, we computed the  mean and SD of the bias across the 14 folds where the subject was included in the training set. Across subjects, the yaw means ranged from $-5.4^\circ$ to $5.4^\circ$ (SDs from $0.1^\circ$ to $0.3^\circ$). The pitch means ranged from $-2.9^\circ$ to $3.9^\circ$ (SDs from $0.1^\circ$ to $0.3^\circ$). The mean and SD for each subject is provided in Table A2 in the supplementary materials. The small SD indicates that $\beta_i$ is stable across folds, implying that $\beta_i$ learns a consistent estimate rather than meaningless random value.
	\section{Usability}
	Our proposed SGPC is demonstrated to reduce estimation error significantly by calibrating on a few images while the user is gazing at a single point. In practice, a calibration procedure can be instructing the subject to look at one target, e.g., the camera, while moving his/her head. Our proposed calibration method is computationally cheap.
	
	For PCCR-based eye trackers, the most commonly used calibration method uses multiple calibration points \cite{guestrin2006general}. This method instructs the user to gaze at different targets sequentially (typical five or nine targets). Compared to it, our proposed SGPC has two major advantages: first, SGPC is simpler and more time efficient as it only needs the user to gaze at one target. Second, it is more widely applicable. In some situations, e.g. screen-free applications, it is difficult to provide multiple targets for MGPC. However, with SGPC, the camera is always a visible calibration point. Although there exists a performance gap between SGPC and MGPC, we believe that SGPC provides a practically useful way to improve accuracy.
	
	\section{Conclusions}
	We proposed a novel gaze decomposition method for appearance-based gaze estimation. This method estimates the person-dependent bias and the gaze of visual axis separately, which improved single gaze point calibration. We conducted experiments on the MPIIGaze, the EYEDIAP and the ColumbiaGaze datasets. Our results indicated that as a subject-independent estimator, the proposed method outperformed the state-of-the-art methods on the MPIIGaze and EYEDIAP datasets. The proposed single gaze point calibration reduced the error significantly while only requiring a few samples (about $9$) looking at one point for calibration and cheap computation. It was also demonstrated to be quite robust to location of calibration points, where the best performance is achieved when calibrated at the center of the gaze range. Our results also demonstrate that an accurate modeling of the data can yield a significant improvement.
	

\input{paper_v1.0.bbl}
	\input{supp}
\end{document}

%% file: supp.tex
	\newpage
	\appendix
	\section{Appendix}
	\begin{table}[b]
		\begin{center}
			\scalebox{0.7}{
				\begin{tabular}{c|c|ccccccc|c}
					\cline{1-10}
					\multirow{2}*{Method}&\multirow{2}*{Vertical}&\multicolumn{7}{c|}{Horizontal}&\multirow{2}*{Average}\\
					\cline{3-9}
					&&$-15^\circ$&$-10^\circ$&$-5^\circ$&$0^\circ$&$5^\circ$&$10^\circ$&$15^\circ$\\
					\cline{1-10}
					\multirow{2}*{LP~\cite{linden2018appearance}}&$0^\circ$&5.2&5.8&5.6&5.2&5.1&5.5&6.3&\multirow{2}*{5.5}\\
					&$-10^\circ$&5.4&5.7&5.5&5.2&5.1&5.2&5.7&\\
					\cline{1-10}
					\multirow{2}*{DF~\cite{liu2018differential}}&$0^\circ$&5.9&5.4&5.1&5.1&5.1&5.3&5.9&\multirow{2}*{5.3}\\
					&$-10^\circ$&5.5&5.1&5.0&5.0&5.0&5.2&5.7&\\
					\hline
					\hline
					\multirow{2}*{\tabincell{c}{Ours\\(w/o GD)}}&$0^\circ$&5.5&5.1&4.9&5.0&5.0&5.2&5.6&\multirow{2}*{5.2}\\
					&$-10^\circ$&5.5&5.1&4.9&4.8&4.9&5.0&5.5&\\
					\cline{1-10}
					\multirow{2}*{Ours}&$0^\circ$&\textbf{5.3}&\textbf{4.7}&\textbf{4.5}&\textbf{4.5}&\textbf{4.5}&\textbf{4.7}&\textbf{5.3}&\multirow{2}*{\textbf{4.7}}\\
					&$-10^\circ$&\textbf{4.9}&\textbf{4.5}&\textbf{4.4}&\textbf{4.4}&\textbf{4.5}&\textbf{4.6}&\textbf{5.1}&\\
					\cline{1-10}
					\multicolumn{10}{l}{\tabincell{l}{\text{*} The calibration-free gaze estimation error of our proposed method with\\ gaze decomposition is $5.5^\circ$.}}\\
				\end{tabular}
			}
		\end{center}
		\caption{Mean Angular Error ($^\circ$) of Calibrating at Different Gaze Targets on the ColumbiaGaze Dataset.}
		\label{tab:ap_error}
	\end{table}
	\begin{table*}[!t]
		\begin{center}
			\begin{tabular}{c|c|cc|cc|cc|cc|cc|cc|cc|cc}
				\cline{1-18}
				{}& {}& \multicolumn{2}{c}{P1}& \multicolumn{2}{c}{P2}& \multicolumn{2}{c}{P3}& \multicolumn{2}{c}{P4}& \multicolumn{2}{c}{P5}& \multicolumn{2}{c}{P6}& \multicolumn{2}{c}{P7}& \multicolumn{2}{c}{P8}\\
				\cline{1-18}
				\multirow{2}*{Yaw}& Mean($^\circ$)&2.8&-2.8& 1.2&-1.2&0.3&-0.4&0.2&-0.2&-1.8&1.9&-4.1&4.0&-0.7&0.7&1.4&-1.5\\
				{}&SD($^\circ$)&0.2&0.3& 0.3&0.1&0.1&0.2&0.1&0.1&0.1&0.1&0.2&0.1&0.1&0.1&0.1&0.2\\
				\cline{1-18}
				\multirow{2}*{Pitch}&Mean($^\circ$)&0.0&0.0& -0.8&-0.8&3.9&3.9&-1.7&-1.8&0.1&0.1&-0.5&-0.4&-1.5&-1.6&0.3&0.3\\
				{}&SD($^\circ$)&0.2&0.2& 0.3&0.3&0.3&0.3&0.2&0.2&0.2&0.2&0.2&0.2&0.2&0.2&0.2&0.2\\
				\cline{1-18}
				\multicolumn{2}{c|}{}& \multicolumn{2}{c}{P9}& \multicolumn{2}{c}{P10}& \multicolumn{2}{c}{P11}& \multicolumn{2}{c}{P12}&\multicolumn{2}{c}{P13}&\multicolumn{2}{c}{P14}&\multicolumn{2}{c}{P15}&{}&{}\\
				
				\cline{1-18}
				\multirow{2}*{Yaw}& Mean($^\circ$)&-3.5&3.4& -0.7&0.9&-0.3&0.2&0.9&-0.9&-0.4&0.3&1.9&-1.8&5.4&-5.4&{}&{}\\
				{}&SD($^\circ$)&0.1&0.1& 0.3&0.2&0.1&0.2&0.2&0.2&0.1&0.2&0.1&0.1&0.1&0.1&{}&{}\\
				\cline{1-18}
				\multirow{2}*{Pitch}&Mean($^\circ$)&-0.6&-0.7& 1.7&1.8&-2.9&-2.9&1.8&1.7&0.2&0.2&-2.8&-2.8&2.9&2.9&{}&{}\\
				{}&SD($^\circ$)&0.2&0.2& 0.3&0.3&0.2&0.3&0.3&0.3&0.1&0.2&0.2&0.2&0.2&0.2&{}&{}\\
				\bottomrule
				\multicolumn{18}{l}{\small{\text{*} For each subject, the left column corresponds to the non-flipped images and the right column corresponds to the horizontally-flipped images.}}\\
				
			\end{tabular}
		\end{center}
		\caption{Mean and Standard Deviation (SD) of the Learned Bias $\hat{b}$ for Each Subject in Training on the MPIIGaze dataset.}
		\label{tab:ap_bias}
	\end{table*}
	\subsection{Cross-Dataset Evaluation}
		For cross-dataset evaluation, we trained on the MPIIGaze dataset~\cite{zhang2015appearance} and tested on the ColumbiaGaze dataset~\cite{smith2013gaze}. For each subject, we calibrated at the five images that correspond to the same gaze target ($T=1, S=5$), and tested on the remaining 65 images. We compared our proposed method with the fine-tuning the latent parameters method (LP) in~\cite{linden2018appearance} and the differential method (DF) in~\cite{liu2018differential}, where they were re-implemented using our architecture. We omitted the comparison with fine-tuning the last fully-connected layer and the adaptation methods since their performance was not good for single gaze target calibration in our previous experiments and in theory. 
		
		In Table~\ref{tab:ap_error}, we present the mean angular errors when calibrating at different gaze targets. Our proposed gaze decomposition (GD) performed the best. On average, it outperformed LP by $0.8^\circ$ ($14.5\%$) and DF by $0.6^\circ$ ($11.3\%$).
	
	\subsection{Evaluation of the Learned Bias}
	We evaluated whether the learned biases, i.e., $\hat{b}_i$ in Eq.~\eqref{train_t} in the main manuscript, were consistent for the same subject using the data from leave-one-subject-out (15 fold) cross-validation on MPIIGaze. 
	
	The mean and SD of each subject are shown in Table~\ref{tab:ap_bias}. Across subjects, the yaw means ranged from $-5.4^\circ$ to $5.4^\circ$ (SDs from $0.1^\circ$ to $0.3^\circ$). The pitch means ranged from $-2.9^\circ$ to $3.9^\circ$ (SDs from $0.1^\circ$ to $0.3^\circ$). We compared the intra-subject variance computed from the 14 folds where the subject was in the training set with the inter-subject variance computed from the means of the estimated biases. For yaw, the average intra-subject variance was $0.03$ $\mathrm{deg}^2$ in comparison to the inter-subject variance of $5.40$ $\mathrm{deg}^2$. For pitch, the variances were $0.05$ $\mathrm{deg}^2$ and $3.66$ $\mathrm{deg}^2$. The intra-subject variance was a small percentage ($0.56\%$-$1.4\%$) of the inter-subject variance, indicating that the bias is learned consistently and reliably during training.